# A Vertical Federated Learning Method For Multi-Institutional Credit Scoring: MICS

Yusuf EFE
*Electrical Engineering Department*
Bogazici University
Istanbul, Turkey
yusufefe@utexas.edu

*Abstract*— As more and more companies store their customers' data; various information of a person is distributed among numerous companies' databases. Different industrial sectors carry distinct features about the same customers. Also, different companies within the same industrial sector carry similar kinds of data about the customers with different data representations. Cooperation between companies from different industrial sectors, called vertical cooperation, and between the companies within the same sector, called horizontal cooperation, can lead to more accurate machine learning models and better estimations in tasks such as credit scoring. However, data privacy regulations and compatibility issues for different data representations are huge obstacles to cooperative model training. By proposing the training framework MICS and experimentation on several numerical data sets, we showed that companies would have an incentive to cooperate with other companies from their sector and with other industrial sectors to jointly train more robust and accurate global models without explicitly sharing their customers' private data.

## I. INTRODUCTION

By the advances in technology, new industries, that are based on data-driven approaches, are emerging. Also, the number of competing companies within the same industry is increasing for most of the industries. Consequently, numerous companies from different industries hold distinct aspects of information about the same customer. For example, while a bank holds mostly financial features about its customers, a telecommunication company stores information about their customers' internet, call, and messaging usage behaviors. To explain the distribution of data among different industries and companies, the phrases "vertical data split" and "horizontal data split" are frequently used in the remaining reading. One should clarify these phrases first:

**Vertical Data Split:** Different industries hold different aspects of a customer's information. In the data set modeling of this situation, features are vertically split into different industries. In Fig. 1, it is illustrated that features a, b, and c belong to the telecommunication industry and features d, e, and f belong to the bank industry.

**Horizontal Data Split:** There are competitors in each industry. While different companies within the same industry store similar kinds of data for their customers, their feature names and units are different. For example, while Telecom A stores weekly internet usage in the unit of Megabyte, Telecom B may store monthly internet usage in the unit of Gigabyte. As a result, they cannot be interpreted as a single data source since the data cannot be revealed explicitly due to data privacy restrictions. In Fig. 2, customers represented by the first 1/3 of the rows are affiliated with Telecom A company.

Fig. 1. Vertical Data Split   Fig. 2. Horizontal Data Split

For the tasks such as credit scoring or recommendation applications, making use of all available data of a customer from various companies would be the most effective scenario to achieve the highest success possible. However, there are mainly two obstacles that deter one from using all the available data:

- Due to data privacy concerns, companies cannot explicitly share their customers' data with other parties. As a result, all the available data cannot be centralized.
- Different companies within the same industry hold similar data but with different data representations. It leads to compatibility issues when being tried to use in a single model.

While existing methods have made significant progress in reaching high accuracies using decentralized data via vertical federated learning methods, they suffer from two limitations. Firstly, most of them are using Homomorphic Encryption [4, 5, 6, 7, 8] to address the data privacy concerns. However, it leads to huge communication costs and makes the training procedure ineffective. Secondly, most of these systems cannot involve more than one company from the same industry due to compatibility issues arising from different data representations. In the proposed framework, MICS, we address both problems and prove that each company in our framework has an incentive to join collaborative training with other companies within their industry and with other industries when compared to the case that they use their own data only.

In our method, as the first step of encryption, we performed a PCA Dimensionality Reduction [3,9] and Random Gaussian Projection [10, 11, 12] to each company's data. After that, each company's data within the same industry is concatenated with a one-hot encoded vector indicating the company, the features belong to. Next, these data are collected in an intermediatory industry server i.e., Banks Server. Following that, these data are forward propagated in the trainable variational encoder neural network called "Banks' Encoder" and the embedding vector representing bank-related information of the customer is produced. Well-known method Variational Autoencoder's methodology [13] is used in the implementation of industrial encoders. During training, the encoder network produced mean vector, $\mu$, and standard deviation vector, $\sigma$. After that, z is sampled as given in Eq. 1 to be fed to the global neural network model as input.

$$\mathbf{z}^{(i,l)} = \mathbf{\mu}^{(i)} + \mathbf{\sigma}^{(i)} \odot \mathbf{\epsilon}^{(l)} \text{ and } \mathbf{\epsilon}^{(l)} \sim \mathcal{N}(0, \mathbf{I}) \qquad (1)$$

So that, a continuous and robust representation of the industrial data in the embedding vector space is achieved. The same procedure simultaneously takes place for the customer's data from the other industries. The resulting embedding vectors are concatenated and fed into the global neural network as the input. The output of this network gives the prediction of the overall model.

## II. RELATED WORK

To review the prior work on this domain, the concepts of Federated Learning [14] and Privacy-Preserving Learning should be analyzed:

**Federated Learning:** It is a machine learning technique that trains an algorithm across multiple decentralized devices holding local data samples, without exchanging them. According to how the data are split across devices, federated learning can be split into two categories as horizontal federated learning and vertical federated learning [15]. Horizontal federated learning, or sample-based federated learning, is introduced in the scenarios that data sets share the same feature space but different in sample. For example, in an image classification task, different edge devices store different image samples, and each edge device could train the model using its own data samples and produce the model parameters' gradients to share with the server. As for Vertical Federated Learning, it is applicable when different edge devices share the same sample space, but they differ in feature space. For example, while a bank stores cash flow features about a customer, a retail company stores the shopping routines of the same customer. Vertical Federated Learning enables utilizing all the features of the same customer across different data holders.

**Privacy-Preserving Learning:** Data privacy is a significant issue. Various privacy-preserving techniques have been developed to allow multiple data holders to collaboratively train Machine Learning models without sharing their private data explicitly. Most of the methods utilize either cryptographic approaches or the concept of differential privacy (DP). Homomorphic Encryption (HME) is the most widely used cryptographic method in Vertical Federated Learning. It enables making computation operations such as addition and multiplication on encrypted data [1].

Utilizing the extension protocols on Homomorphic Encryption, comparison of two encrypted values and performing secure multiplications become possible. As for DP, it is concerned with the output of computations on a data set that can leak information of the individual entries from the data set. Utilizing the randomness in DP, leakage of such entry details can be prevented.

Recent work on Vertical Federated Learning uses either HME [4, 5, 6, 7, 8] or DP [16, 17, 18] mostly. Several other works [19,20,21,22,23,24,25,26] use different encryption mechanisms. Nearly all these works either have high communication costs due to encryption mechanism or lacking the ability to include different data representations within the same feature group (industry) in one global model training. In the work developed by Chen et al. [27]., there is a similar approach to MICS but we differentiate with two factors: Firstly, we enable different data representations within the same feature group (industry) to jointly train one global model. Secondly, there are two steps of encryption in MICS.

## III. METHOD

### A. Problem Definition

Consider a set of M industries: $\mathcal{M} \coloneqq 1, \dots, M$. For example, industry 1 may be the telecommunication sector and industry 2 may be the banking sector. Each industry k consists of $N_k$ competing companies. In the problem definition, it is assumed that each customer is affiliated with a company from each industry. As a result, each company, the customer is affiliated with, holds different aspects of the customer's information. A data set of Z customers, $\{x_i, y_i\}$, is maintained by all the companies' data about their customers, and the number of companies, T, is equal to:

$$T = \sum_{k=1}^{M} N_k \qquad (2)$$

In an industry, similar kinds of customer data are being stored by the companies, but their representations differ. For example, while Telecom Company A holds internet feature as weekly GB usage, Telecom Company B may hold monthly internet usage in the unit of MB. Since there is no explicit reach to these features, a method is needed to tackle the compatibility issue.

### B. MICS Method

All companies' data from the same industry are reduced to the same fixed size vectors via the PCA algorithm. PCA fit is applied to each company separately. After this step, to indicate which company the data is coming from, the one-hot encoded vector representing the company is concatenated with the PCA fitted data.

To mathematically represent the data, denote $\mathbf{x}_{n,m,k}$ as the feature vector, $n^{th}$ company from $m^{th}$ industry holds about customer k. To deal with the compatibility issues, each company's data in the same industry are PCA fitted separately to the same sized vectors. After this step, all companies in a particular industry hold the same length feature vectors about their customers. However, the meanings of these feature vectors still differ for different companies as explained in the previous example. After that, they are one-hot encoded to represent which company the data belongs to. Although all companies in an industry hold the same length feature vectors about their customers, their meanings completely differ and by concatenating one-hot encoded vector to the original feature vector, we tried to let neural network learn different representations as a function of the data source utilizing one-hot encoded vectors.

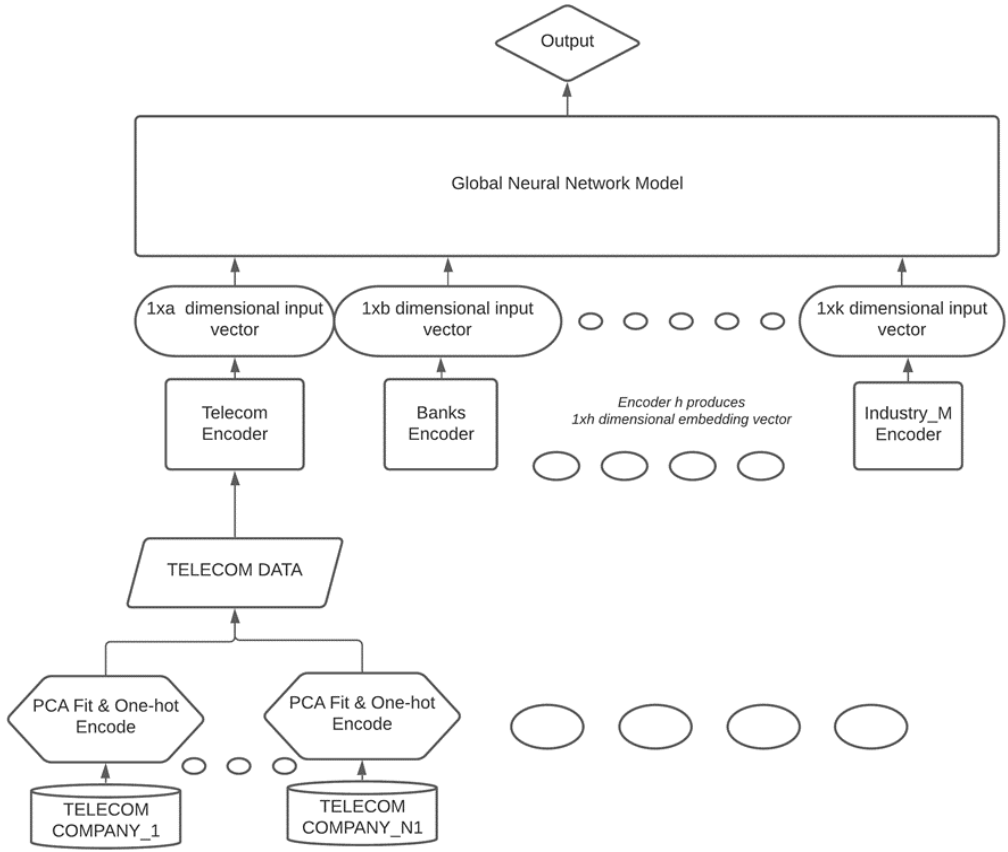

Fig. 3. MICS Architecture

The transformed version of the company data is denoted as $\mathbf{x'}_{m,k}$. Here, the subscript n is removed since all the data from the same industry are compatible after the transformations are applied. At this step, all the companies' data from the same industry are collected in the industry server. This procedure takes place in all industry servers simultaneously. Following this, according to customer IDs, industry data about the customers are aligned in a way that the $k^{th}$ row in each industry server data represents the same customer. By forward propagating the data in each industry encoder, industry-embedding vectors are produced for the customers. For example, the industry-1 embedding vector gives encrypted information about the customer in the Telecommunication sector. After that, all industry-specific embedding vectors are concatenated and this resulting vector is fed to the global neural networks as the global encrypted input which holds information about the customer data from all industries she is affiliated with. Finally, this global input is forward propagated in the global neural network and the model's prediction about the problem is produced.

During the training of the model, backpropagation takes place in the global model first and the global model is updated. After that, gradient vectors for each industry-embedding are produced separately and each industry-specific encoder's model parameters are updated using the gradient values of the embedding vector it produced. The architecture of the model is given in Fig. 3.

### C. Key Advantages of MICS Framework:

**Ability to use all available data from various companies to train a single global model:** MICS can use not only different industries' data but also each company's data within the same industry although they have distinct data representations. To stress the importance of this ability, let us analyze a concrete example. Imagine that there are five different industries and four competing companies in each industry. For the sake of simplicity, assume that each of these 20 companies holds k-length feature vectors about their customers. Also, assume that m customers are split as m/4, m/4, m/4, and m/4 to four competing companies in each industry. Now, let us have a look at different scenarios and how many rows and features can be used in these different scenarios during training of the global model:

- **MICS:** MICS can utilize all the available data, meaning that 5k-length feature vector for m customers. It amounts to 5km feature points.

- **Cooperation only within the industry:** In this scenario, k-length feature vectors are utilized for all m customers. It amounts to 1km feature points.
- **One company cooperates with all companies from only other industries:** In this case, 5k-length feature vectors are utilized for m/4 customers data. It amounts to 1.25km feature points.
- **One company from each industry establishes a 5-company coalition:** In this case, they can only use their common customers to train the model which significantly decreases the number of available customers. In this case, assuming the customers are split IID among companies within each industry, 5k-length feature vectors are utilized for m*(1/4)5 customers. It amounts to 5km/1024 feature points.
- **No collaboration– (each company uses its own data only):** In this scenario, each company trains its own model using k-length feature vectors for m/4 customers. It amounts to km/4 feature points.

**Data Encryption steps match the regulations:** In the MICS framework, there are mainly two encryption steps: (PCA fitting + Random Gaussian Projection) and industry-specific encoder model processing. PCA + Random Gaussian Projection's power of encryption is weaker compared to encoder neural network's encryption power. However, it coheres with the real-life needs of data privacy standards. In most countries, governments' industry-specific regulation maker institutions have a right to centralize all the companies' data of their industry in partially encrypted form. Thus, they can behave as intermediary industry servers to prevent private data leakage. Before concatenating all embedding vectors, they are encrypted via the encoder networks. Thus, the second centralization of the data does not lead to any data privacy-related problem.

## IV. EXPERIMENTS

In this section, we evaluated our method on 3 different publicly available datasets: House Prices Dataset and Red-White Wine Quality Dataset [2]. These data sets consist of numeric features only. We transformed the labels into binary classification tasks to be able to gain a wider range of accuracies for different settings. To represent multiple industries from a central data set, vertical splitting the dataset is adequate. However, horizontal splitting is not enough to represent different companies in an industry. We needed to change the data representation while keeping the prediction power of the data the same to mimic the real-life scenario. For this purpose, Random Gaussian Projection Method [4] is used. To give a concrete example, let's assume that we have a centralized data set with m rows and n features. For example, the "two industries and three companies" case corresponds to the situation that n/2 features belong to the first industry and the other n/2 features belong to the second industry. As for rows, at each industry, m rows are distributed among three companies equally (m/3, m/3, and m/3). So, in each industry, we need to create different company representations. In this case, we applied Random Projection to each of these 6 companies with different random seeds to create distinct data representations while preserving their predictive power.

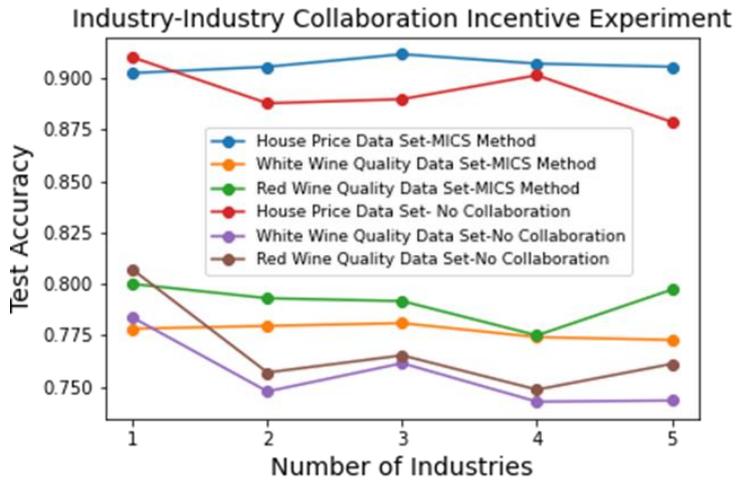

Fig. 4: Industry-Industry Collaboration Incentive Experimentation

### A. Collaboration Among Industries Incentive Experiment

In this experiment, the incentive of collaboration between industries is investigated. The company count is kept as 1 during the experimentation. In one scenario, the MICS framework is used, and the number of industries is varied from 1 to 5 as the independent parameter. For the alternative scenario, industries used only their own features to train and test the model. Comparisons are given in Fig. 4.

Industries have an incentive to establish collaborations with other industries to achieve higher accuracy in their tasks. For all three data sets, the "MICS Method" scenario outperforms the "No Collaboration" scenario as the number of industries increases.

### B. Collaboration Among the Companies within the Same Industry Incentive Experiment

In this experiment, the incentive of a company to build collaborations with other companies from the same industry is investigated. Industry number is kept as 1 during the experimentation and the effect of company number is analyzed. In one scenario, the MICS framework is used, and the number of company representations within the industry is varied from 1 to 5 as the independent parameter. In the alternative scenario, each company used its own data only to train and test the model. Comparisons of these two scenarios for three data sets are given in Fig. 5.

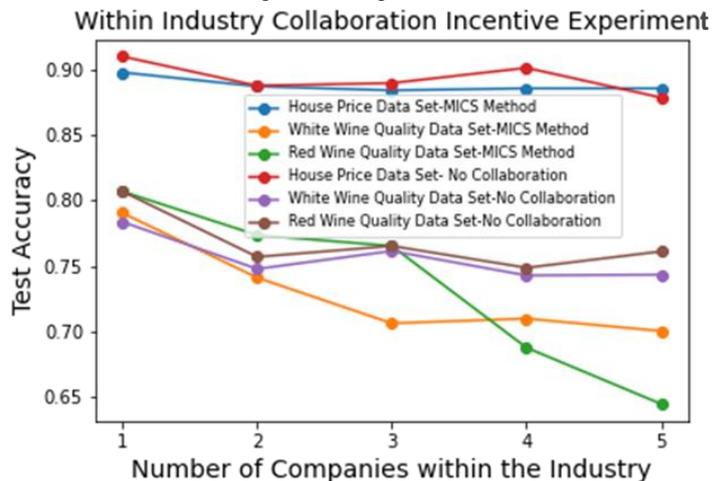

Fig 5: Collaboration within Industry Incentive Experimentation

According to the experimentation results, the "MICS Method" and the "No Collaboration" scenario did not differ significantly for the "one industry, varying company number" case. Thus, companies would not have an incentive to build collaborations with other companies from their industry when no other industry is present. To see their behavior when other industries present, we can proceed to *Experiment C*.

### C. Comparison of MICS and No Collaboration Scenarios for Several Cases

In this experimentation, "MICS Method" and "No Collaboration" scenarios are compared for several cases by varying industry count and company count. Results show that any company would have an incentive to join MICS by establishing collaborations with other industries and with the companies within their industry. While "within industry collaboration" did not make much difference for the case of one industry, it created a significant impact for the cases when industry count is greater than one.

Surprisingly, the strong side of the MICS method is that its success is not affected by the number of industries. Regardless of the number of industries, the data is vertically split into, MICS can reach similar accuracy scores. However, the vertical collaboration did not create much difference other than letting us use all the available data to train one common global model. Thus, a better way of "collaboration within the industry" could be further investigated.

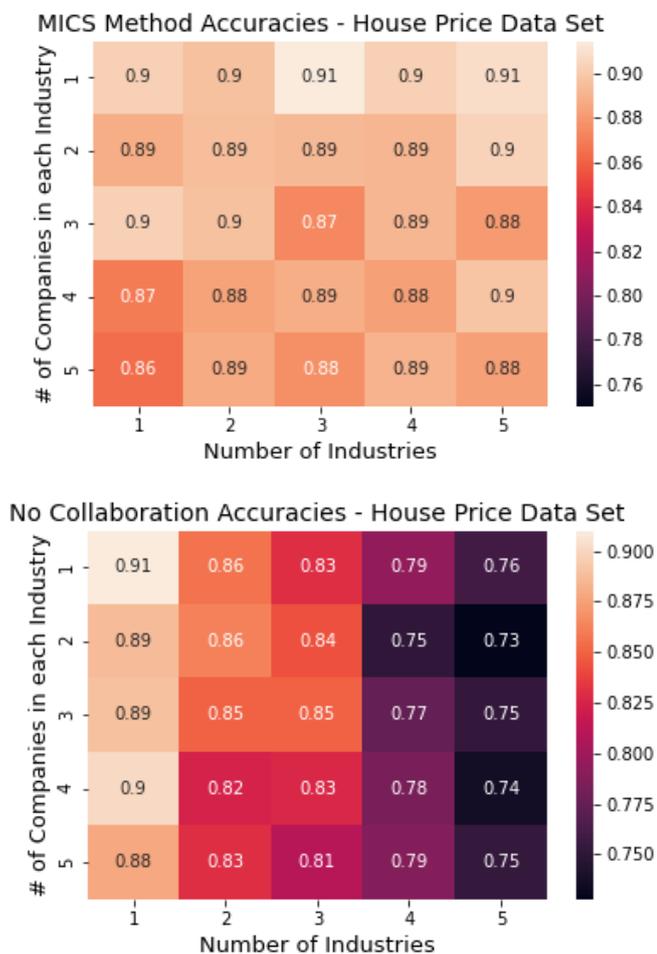

Fig. 6. "MICS" vs "No Collaboration" Case Comparisons for Various Combinations of Industry Count and Company Count

## V. CONCLUSION

In this work, we tackle the obstacles of data privacy concerns and data compatibility issues to enable multiple institutions to jointly train one global model without revealing their customers' data for tasks such as credit scoring. We propose vertically splitting data into different industries, horizontally splitting & transforming industry data into different companies, and training one separate variational encoder for each industry. Using variational encoders, smoothly continuous representation of input data in embedding vector space is provided. Also, concatenating one-hot encoded vectors to original inputs, we enable all companies within an industry to jointly train one common variational encoder. After generating embedding vectors for each industry, the global input vector is derived via concatenation of industry-embedding vectors. Finally, this global input is forward propagated in the global neural network and the output is generated.

We showed that in multi-institutional data-based tasks such as credit scoring, our framework MICS can achieve better accuracy scores compared to non-cooperative alternatives without violating any data privacy concerns. Also, it has low communication costs compared to the previous methods built on Homomorphic Encryption.

As the future work in this domain, reverse engineering threats for private data leakage and how to make the system more robust to such kind of adversarial attacks can be investigated.

## VI. ACKNOWLEDGMENT

I thank Dr. Mustafa Kirac and Prof. Dr. Murat Saraclar for their helpful comments and suggestions.